\documentclass{article} 
\usepackage{iclr2023_conference,times}

\usepackage{amsmath,amsfonts,bm}

\def\eqref#1{equation~\ref{#1}}

\def\1{\bm{1}}

\DeclareMathAlphabet{\mathsfit}{\encodingdefault}{\sfdefault}{m}{sl}
\SetMathAlphabet{\mathsfit}{bold}{\encodingdefault}{\sfdefault}{bx}{n}

\usepackage{hyperref}
\usepackage{url}
\usepackage{graphicx}

\title{Global Flood Prediction: \\A Multimodal Machine Learning Approach}

\author{Cynthia Zeng\\ 
Massachusetts Institute of Technology\\
\texttt{czeng12@mit.edu} \\
\And
Dimitris Bertsimas\\
Massachusetts Institute of Technology \\
\texttt{dbertsim@mit.edu}
}

\iclrfinalcopy 
\begin{document}

\maketitle

\begin{abstract}
Flooding is one of the most destructive and costly natural disasters, and climate changes would further increase risks globally. 
This work presents a novel multimodal machine learning approach for multi-year global flood risk prediction, combining geographical information and historical natural disaster dataset. 
Our multimodal framework employs state-of-the-art processing techniques to extract embeddings from each data modality, including text-based geographical data and tabular-based time-series data. 
Experiments demonstrate that a multimodal approach, that is combining text and statistical data, outperforms a single-modality approach.
Our most advanced architecture, employing embeddings extracted using transfer learning upon DistilBert model, achieves 75\%-77\% ROCAUC score in predicting the next 1-5 year flooding event in historically flooded locations. 
This work demonstrates the potentials of using machine learning for long-term planning in natural disaster management. 
\end{abstract}

\section{Introduction}

A disastrous flood in 2022 left one third of the land in Pakistan underwater for over four months, affecting 33 million people in the country and causing over 30 billion US dollars of damage \citep{united_nations_2023}. Globally, floods cost billions of dollars each year and inflict massive damage to human life, infrastructure, agriculture, and industrial activities. 
Most concerningly, studies suggest climate change impacts lead to drastically increasing flooding risks globally in both frequency and scale \citep{wing_inequitable_2022, hirabayashi_global_2013}. Therefore, it is crucial to develop both short-term and long-term predictions for flood events to mitigate damage.

Most established models for flood prediction use physical models to simulate hydrological dynamics. 
\citet{kauffeldt_technical_2016} provides a technical review of large-scale hydrodynamical models employed in various continents. 
The most advanced models take into consideration terrain data, water flow data, river networks \citep{sampson_high}. 
To combine insights from individual models and reduce errors, most forecasting agencies, such as the pan-European Flood Awareness System (EFAS), employ an ensemble of predictions across many individual hydrological models to produce probabilistic forecasts \citep{thielen2009european}.

Physical models dominate short-term flood prediction space; however, they lack forecasting capabilities for a longer horizon due to escalating simulation errors. 
To address this need, machine learning can emerge as a powerful tool to offer a predictive perspective. \cite{mosavi_flood_2018} provides an extensive literature review on the recent ML approaches.
Most early works of machine learning approaches are based on a single modality of data, such as rainfall and water level data \citep{sajedi-hosseini_novel_2018, choubin_precipitation_2018, elsafi_artificial_2014}, or remote-sensing dataset such as satellite and radars to capture real-time high resolution rain gauges \citep{kim_quantitative_2001, sampson_high}. Multimodal machine learning, referring to models that employ more than one modality of data such as tabular, imagery, text, or other formats, have been recently applied for flood detection purposes. For instance, \citet{twitter_flood} combines hydrological information with twitter data to detect and monitor flood. 

This work presents a multimodal machine learning approach combining for global multi-year flood prediction.
 To the best of our knowledge, this is the first machine learning flood prediction model at the global scale and on a multi-year horizon. In addition, it is the first time text-based data has been applied to flood prediction.
Our main contributions are three-fold: 
\begin{enumerate}
    \item A novel multimodal framework to incorporate text-based geographical information to complement time-series statistical features for global flood prediction. We employ state-of-the art large natural language processing techniques, including fine-tuning and transfer learning on pre-trained BERT models.  
    
    \item  Our experiments show strong results for multi-year flood risk forecasting, with the strongest model achieving 75\%-77\% ROCAUC score in the next 1-5 year flooding prediction. In addition, we show that multimodal models, combining text with statistical data, outperform single-modal models using only statistical data. 
    
    \item Our framework can be generalised to other natural disaster forecasting tasks such as the wildfires, earthquakes, droughts, and extreme weather events. Thus, this works suggests a promising direction in long-term preparation for natural disaster management. 
    
     
\end{enumerate}

\section{Data}
\paragraph{Historical Flood Data.}
We use the Geocoded Disasters (GDIS) dataset, which includes geocoded information on 9,924 unique natural disasters occurred globally between 1960 and 2018 \citep{rosvold_geocoded_2021}. In addition, we linked this dataset with the EM-DAT dataset to add additional economic information such as damage estimation \citep{emdat}. 
In this project, we restrict forecasting locations to those with historical flooding event. We use the date, latitude, longitude, location (given as the name of the location), and if available, damage cost from this dataset. 
We divide the earth into 1$^{\circ}$ by 1$^{\circ}$ grid, corresponding to about 100km by 100km squares. Using the latitude and longitude information, we compute a `grid id' for each natural disaster from the GDIS dataset. Overall, there are 2852 unique grid locations in the dataset with a recorded historical natural disaster.

\begin{figure}[h]
    \centering
    \includegraphics[width=0.8\textwidth]{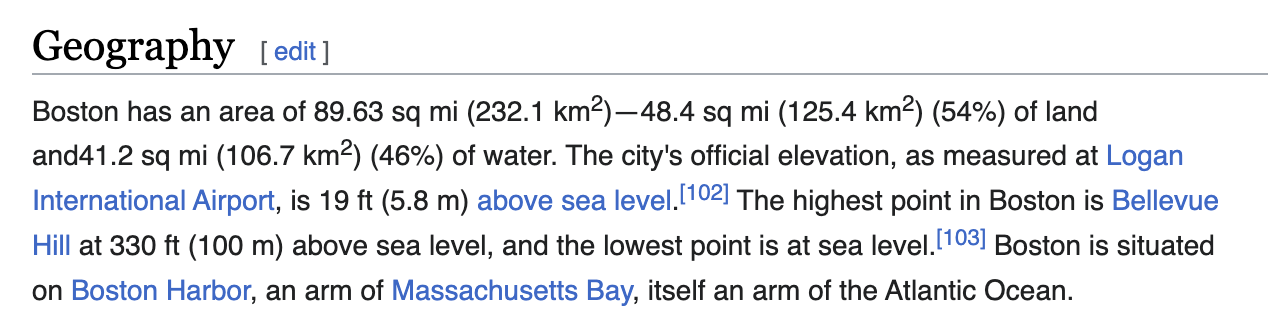}
    \caption{Example `Geography' section of the Boston Wikipedia page.}
    \label{fig:boston_wiki}
\end{figure}

\paragraph{Geographical Data.}
To incorporate the geographical information of each location, we use open-source Wikipedia website's Geographical section, which contain text-based geographical description of certain areas, as shown in Figure \ref{fig:boston_wiki} as an example for the `Boston' Wikipedia page. 
To obtain the geographical information, we use the `location' data from the GDIS dataset for each grid id, then use the Wikipedia-API to obtain the text from the Geographical section for each location \citep{wikipedia-api}. To deal with the noise in the data, since some locations have different names on Wikipedia, we search over synonyms for each location. For those location Wikipedia pages without Geography section, we use the Summary section. Among 2852 unique grid ids, we collected text-based information for 2775 grid ids, and fill the remainder grid ids as `missing'. 

\section{Methodology}
\begin{figure}[h]
    \centering
    \includegraphics[width=\textwidth]{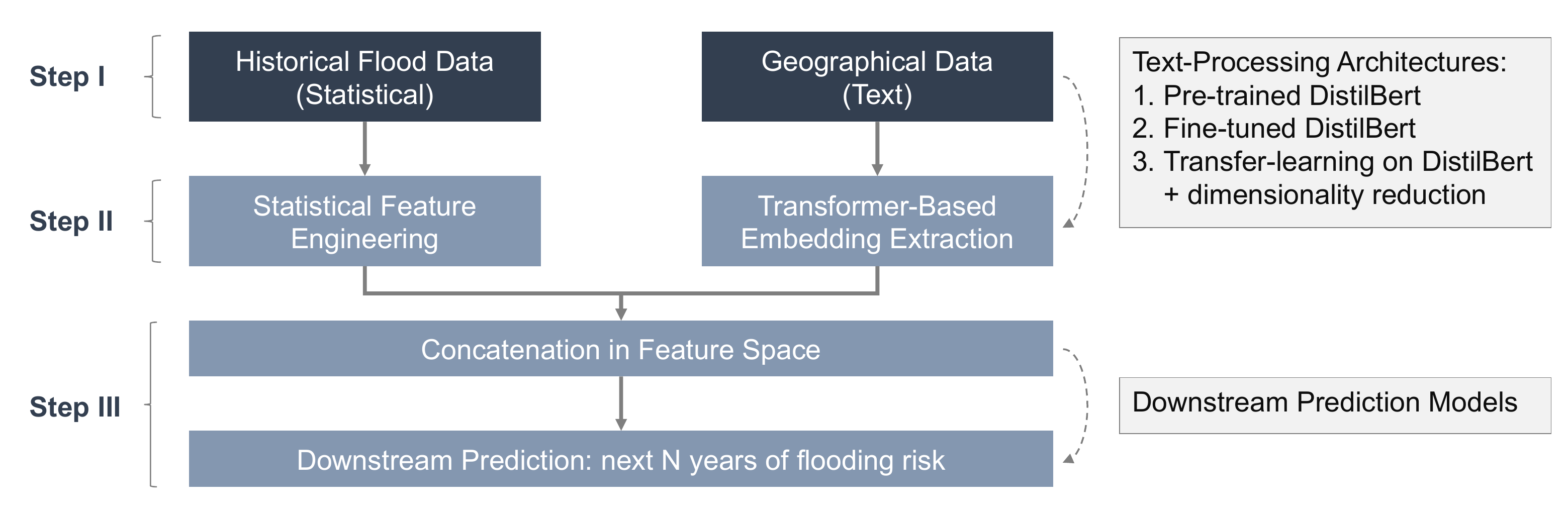}
    \caption{Three-step framework to combine statistical data with text-based data. The transformer-based text data embedding extraction contains three types of architectures.}
    \label{fig:framework}
\end{figure}

The overall goal is to predict next 1 to 5 years of flood risk using a multimodal approach. The framework adopts a three-step approach to combine distinct data formats and sources. Figure \ref{fig:framework} illustrates the overall three-step framework. More details of the training and testing protocol can be found in the Appendix. 

\begin{enumerate}
    \item We gather different sources and modalities of data, which are a) tabular-based historical natural disaster data and b) text-based geographical data from Wikipedia pages. 
    \item We perform feature processing individually for each data modality, and obtain a one-dimensional feature representation (embeddings) respectively. 
    \item We concatenate feature embeddings from different modalities and perform feature sections, before making next-N-year flood event predictions using gradient boosted tree (XGBoost) models for binary classification task. Prediction target 1 indicates a flood in the next N years, 0 otherwise. 
\end{enumerate}

\subsection{Statistical Feature Processing}

We use the GDIS dataset to process historical statistics of natural disasters. In particular, for each grid id, we aggregate statistical features into yearly basis uisng only the current year's natural disaster statistics. In particular, we summarize the `count' `binary' and `damage cost' feature during the year for each natural disaster: `flood', `storm', `earthquake', `extreme temperature', `landslide', `volcanic activity', `drought', `mass movement (dry)'. The `damage cost' feature corresponds to the insurance amount claimed by the natural disaster, which is intended as a proxy to reflect the severity of the natural disaster. In total, the statistical features contain 24 features. Additionally, we record the `year' feature as numerical feature. 

\subsection{Text Feature Processing}
For each location, we use the Geography section from the Wikipedia page using the location name. This information is given as text, and each location is associated with a paragraph of geographical information description.   
Under the scope of this work, we experiment with pre-trained large language model DistilBert, a distilled version of the BERT model, which offers good performance whilst faster to train and fine-tune \citep{sanh2019distilbert}. 
The two main challenges are: a) DistilBert model is trained on a large set of generic texts, whilst we would like to adapt it to encode geographical information specifically; b) feature extraction is performed on a token-by-token basis, whilst we require embeddings corresponding to a paragraph of sentences. In summary, we experiment with three distinct architectures. 

\begin{enumerate}
    \item The original DistilBert. As proposed by \citet{li2020sentence}, we use the second last layer of hidden states and taking the average of embedding tokens across from all words in the sentence to obtain the paragraph embedding. 
    \item Fine-tuned version of the DistilBert model. We fine-tune the DistilBertForSequenceClassification model using binary classification labels with 1 indicating the location has more than two historical floods, and 0 indicating the location has less or equal to two historical floods. The motivation is to fine-tune DistilBert embeddings specifically for flood prediction. Then we pool token embeddings by taking the average of the second last layer. 
    \item Transfer learning and dimensionality reduction. We add an additional linear layer of dimension (796, 32) with a sigmoid activation function. The classification labels are the same as in the second approach, and we use the 32 vector as extracted embeddings. During the training process, parameters from the pre-trained  model are frozen, and the training only learns parameters from the linear layer. Similarly as above, we compute paragraph embeddings by taking the average of the 32-vector embeddings for each token. 
\end{enumerate}

\section{Results}

Table \ref{tab:results} contains out-of-sample binary classification performance from various models for the next 1,2,5 year flood prediction horizon on the selected 818 grid locations. In summary, a multimodal approach demonstrates the strongest performance, achieving 70\% - 75\% ROCAUC score. Training and testing sets are randomly selected at 70\% and 30\%, and more details on the training protocols can be found in the Appendix. 
 
We construct a deterministic baseline model which predicts the next N years of flood outcome as the same current year flood outcome. This approach aims to mark previously flooded region as high risk, which is similar to the flood risk mapping procedure employed by agencies such as FEMA. 

Due to high class imbalance, metrics such as ROCAUC and balanced accuracy scores are more objective than accuracy scores in evaluating prediction capabilities. We observe that a single-modality model employing only statistical features outperforms the baseline model by around 35\% in ROCAUC score and around 25\% in balanced accuracy, underperforms the baseline by around 23\% in accuracy score. Among multimodal approaches, the strongest architecture combines statistical features with text features obtained using transfer learning upon DistilBert model. This architecture improves upon the baseline model by around 42\% in ROCAUC score, 25\% in balanced accuracy,and underperforms the baseline by around 13\% in accuracy score. Finally, other multimodal architectures, such as using directly pre-trained DistilBert or finetuned DistilBert does not improve the performance from a single-modality approach.



\begin{table}[h]\label{tab:results}
    \centering
\includegraphics[width=0.8\textwidth]{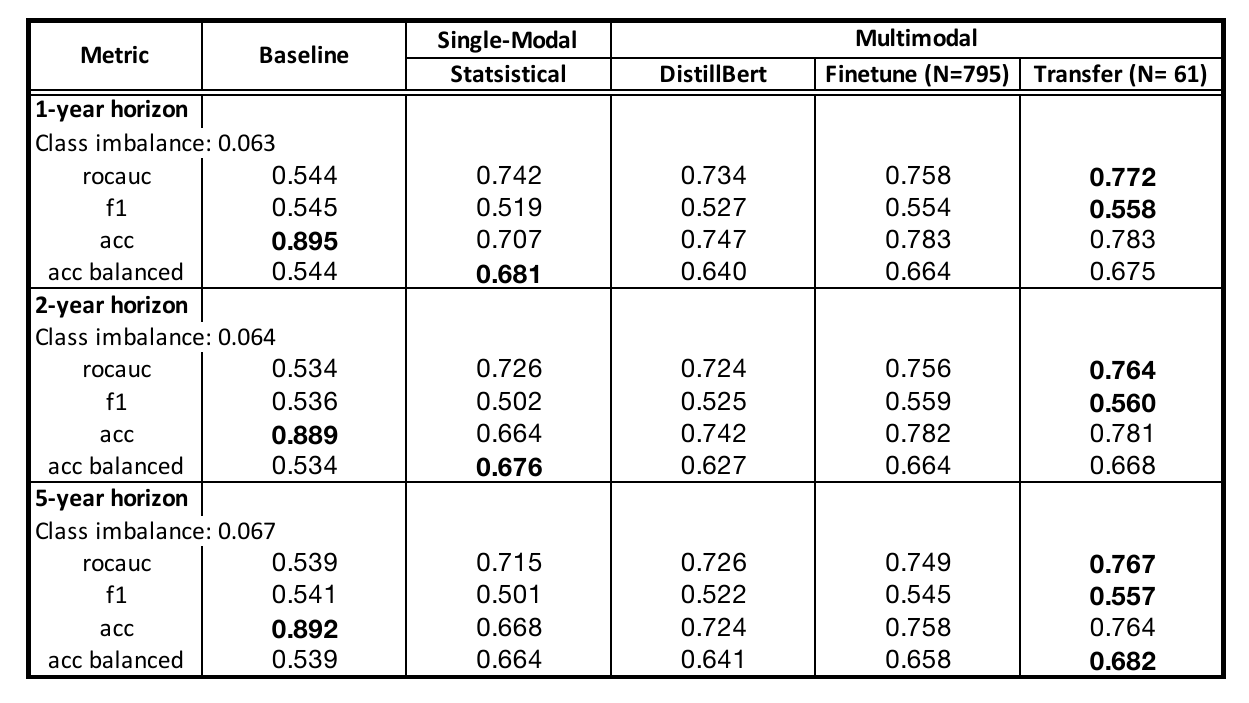}
\caption{Out-of-sample performance for the next 1,2,5 years of flood risk prediction task. 
Baseline model predicts the same outcome as current year outcome. 
Multimodal models employs statistical features and text embeddings extracted using various architectures. 
We record the number of total features employed in each approach given in brackets. 
We report ROCAUC score, accuracy, F1 score, and balanced accuracy.}
\end{table}

\section{Conclusion}

This work presents a multimodal machine learning framework for global flood risk forecasting combining statistical natural disaster dataset with text-based geographical information. 
This work demonstrates strong results for multi-year flood risk forecasting globally, enabling potentials for long-term planning in natural disaster management. 

\newpage 
\bibliography{iclr2023_conference}
\bibliographystyle{iclr2023_conference}

\appendix

\section{Training and Testing Protocol}

In Step II, for the fine-tuning and transfer learning of transformer-based feature extraction models, we split the text dataset (which contains 2852 locations with associated Wikipedia text data) into training and validation set with 70\% randomly selected samples as the training set. Models are trained using SGD with Adam optimiser. Both fine-tuning and transfer learning are trained on 3 epochs.  

In Step III, for the training and testing of the downstream binary classification task of flooding risk, we separate the data into 70\% training and 30\% testing. For each model, we perform 3-fold cross validation on the grid search to perform hyperparameter tuning with AUC score as the scoring metric. we record the following evaluation merics: accuracy, balanced accuracy, ROCAUC score, and F1 score.


The training and fine-tuning of DistilBert models are conducted on Google Colab with 1 GPU computing power. The training and parameter search on classification tasks are conducted using the MIT SuperCloud cluster with 1 GPU computing power \citep{MIT_supercloud}. 

As a remark, due to the rarity of natural disaster occurrence, we face a significant data imbalance challenge: the majority of the grids would not have a flood incidence and, thus, the positive prediction case is less than 0.1\% for the entire dataset. To address this issue, we filter to select grid ids with at least 2 historical flood incidents, and perform prediction tasks on those selected grid ids. This filtering criterion is based on the assumption that some grid locations are not prone to flooding risk. Among 2852 unique grids, 881 grids are selected.

\end{document}